\documentclass[letterpaper]{article} 
\usepackage{aaai2026}  
\usepackage{times}  
\usepackage{helvet}  
\usepackage{courier}  
\usepackage[hyphens]{url}  
\usepackage{graphicx} 
\usepackage{natbib}  
\usepackage{caption} 
\usepackage{hyperref} 
\usepackage{algorithm}
\usepackage{algorithmic}
\usepackage{amsmath,amssymb,amsthm}
\usepackage{booktabs} 
\usepackage{multirow} 
\usepackage{makecell} 
\theoremstyle{plain} 

\usepackage{newfloat}
\usepackage{listings}
\usepackage{subcaption}
\DeclareCaptionStyle{ruled}{labelfont=normalfont,labelsep=colon,strut=off} 
\floatstyle{ruled}
\newfloat{listing}{tb}{lst}{}
\floatname{listing}{Listing}
\title{JustDense: Just using Dense \\instead of Sequence Mixer for Time Series analysis}
\author{
    TaekHyun Park\textsuperscript{\rm 1}\equalcontrib,
    Yongjae Lee\textsuperscript{\rm 1}\equalcontrib,
    Daesan Park\textsuperscript{\rm 1}\equalcontrib,
    Dohee Kim\textsuperscript{\rm 1},
    Hyerim Bae\textsuperscript{\rm 1}\thanks{Corresponding author} \\
}
\affiliations{
    \textsuperscript{\rm 1}Pusan National University, Busan, Korea\\

    \texttt{\{pthpark1, yongzzai1102, ajtks9414, kimdohee, hrbae\}@pusan.ac.kr}

}

\begin{document}

\maketitle

\begin{abstract}
    Sequence and channel mixers, the core mechanism in sequence models, have become the
    \textit{de facto} standard in time-series analysis (TSA).
    However, recent studies have questioned the necessity of complex sequence mixers, such as attention mechanisms,
    demonstrating that simpler architectures can achieve comparable or even superior performance .
    This suggests that the benefits attributed to complex sequence mixers might instead emerge from
    other architectural or optimization factors. 
    Based on this observation, we pose a central question: \textbf{Are common sequence mixers necessary for time-series analysis?}
    Therefore, we propose \textbf{JustDense}, an empirical study that systematically replaces sequence mixers in
    various well-established TSA models with dense layers.
    Grounded in the $MatrixMixer$ framework, JustDense treats any sequence mixer as a mixing matrix and replaces it with a dense layer.
    This substitution isolates the mixing operation, enabling a clear theoretical foundation for understanding its role.
    Therefore, we conducted extensive experiments on 29 benchmarks covering five representative TSA tasks using seven 
    state-of-the-art TSA models to address our research question.
    The results show that replacing sequence mixers with dense layers yields comparable or even superior performance.
    In the cases where dedicated sequence mixers still offer benefits, JustDense challenges the assumption that "deeper and more complex architectures are inherently better" in TSA.
\end{abstract}


\section{Introduction}
Time Series Analysis (TSA), which examines data that varies over time, is essential across various domains, 
such as finance~\cite{capistran2010multi}, logistics~\cite{kim2025long}, and weather forecasting~\cite{bi2023accurate},
driving high-stakes real-world decisions.
In these domains, time series data are utilized for tasks, such as Long and short-term forecasting, missing value imputation, classification, and anomaly detection~\cite{wang2024deep}, 
enabling extraction of meaningful insights and patterns.
Effective TSA is often ensured by accurate prediction of numerical values at certain points in time.
Traditionally, statistical models, such as Auto-Regressive Integrated Moving Average (ARIMA) and Error-Trend-Seasonal (ETS), 
have been employed to capture the sequential nature of time-series data~\cite{sim2023correlation}.
However, sequence models with complex sequence mixers, such as attention mechanism~\cite{vaswani2017transformer}, 
have replaced this trend.

Modern TSA architectures commonly adopt a two-stage pipeline: a sequence mixer that aggregates information across the temporal axis,
followed by a channel mixer that refines representations within each time step.
\cite{hwang2025hydra} proposed the \textit{MatrixMixer} framework, which interprets all existing sequence mixers as an
$L\times L$ mixing matrix, providing a unified perspective for comparison.
Within this framework, state-of-the-art TSA models fall into three families based on the type of sequence mixer:
(1) Attention-based, (2) Toeplitz-based (ex. Convolution), and (3) Semiseparable-based (ex. State Space).
For example, Transformer~\cite{vaswani2017transformer}, PatchTST~\cite{Yuqietal-2023-PatchTST}, and iTransformer~\cite{liu2024itransformerinvertedtransformerseffective}
all include the attention mechanism, while differing in architectural components.
Similarly, the other two families, toeplitz-based and semiseparable-based, include various models that employ the same sequence mixer (e.g., Convolution).

Although complex sequence mixers currently dominates TSA, recent research has begun to question their necessity.
In the absence of sequence mixers, DLinear~\cite{Zeng2022AreTE} employs a single linear layer for forcasting, 
and TSMixer~\cite{chen2023tsmixer} replaces the sequence mixer with lightweight MLP blocks, 
yet both match or even surpass attention-based baselines on long-term time series forecasting benchmarks.
However, these models have not been compared or analyzed in a unified framework to examine the underlying mathematical structure underlies the models
and their influences on properties, such as computational complexity and causality.
Concurrently, \cite{hwang2025hydra} replaced the sequence mixer in the Mamba~\cite{gu2023mamba} architecture,
originally based on semiseparable matrix, with simpler alternatives and demonstrated comparable or improved performance.
Nevertheless, this evidence is restricted to natural language processing and Mamba architecture; thus, it has remained untested in TSA.

These observations suggest that the benefits traditionally attributed to complex sequence mixers may instead emerge from other architectural or optimization factors.
Therefore, the following research question can be posed:
\begin{quote}
    \textbf{RQ. }\textit{Are common sequence mixers truly necessary for time series analysis?}
\end{quote}
To study this question, we propose JustDense, an empirical study that challenges the necessity of common sequence mixers in TSA.
Specifically, sequence mixers— attention-based, toeplitz-based, or semiseparable matrix-based— are replaced with simple dense mixers,
enabling the isolation and evaluation of the intrinsic value of the underlying model architecture.
Experiments are conducted on 29 public benchmarks spanning five representative time series tasks—classification, anomaly detection, imputation, long-term forecasting, and short-term forecasting—against seven baseline TSA architectures.
Our main contributions are summarized as follows:
\begin{itemize}
    \item We generalize recent uncertainty regarding the necessity of conventional sequence mixers and explicitly investigates what drives effective sequence mixing in TSA.
    \item We propose \textbf{JustDense}, a framework grounded in the MatrixMixer that systematically replaces the $L\times L$ mixing matrices of state-of-the-art TSA models with simple dense layers, providing a clear theoretical foundation for isolating architectural effects.
    \item Across 29 benchmarks covering five representative TSA tasks (classification, anomaly detection, imputation, long-term forecasting, and short-term forecasting) and seven baseline TSA architectures,
    we demonstrate that without convectional sequence mixers achieve comparable or superior performance, challenging the assumption that more formal structures are inherently better in TSA.
\end{itemize}

\section{Related Works}
\label{sec:relatedwork}
This section reviews existing studies organized into three paradigms of sequence mixers: Attention-based, Toeplitz-based, and Semiseparable-based.

\paragraph{Attention-based}
Initially developed for natural language processing, 
Attention-based models~\cite{vaswani2017transformer} were rapidly adapted for time series forecasting owing to their proficiency in capturing long-range dependencies. 
To mitigate the quadratic complexity of the original attention mechanism, Informer~\cite{haoyietal2021informer} introduced the ProbSparse attention mechanism.
 Recently, PatchTST and iTransformer achieved state-of-the-art performance by applying patching 
 and channel-independent strategies, demonstrating the continued relevance and adaptability of the Transformer architecture.

\paragraph{Toeplitz-based}
Toeplitz-based models offer a computationally efficient approach to capturing local temporal patterns.
Autoformer~\cite{autoformer} replaced attention with an autocorrelation-based mechanism that captures temporal dependencies using a convolutional approach. 
Recently, PatchMixer~\cite{gong2023patchmixer} utilize Toeplitz-structured depthwise convolutions (DWConv) and pointwise convolutions on patched time series, 
achieving competitive performance with significantly lower computational overhead.
Furthermore, ModernTCN~\cite{luo2024moderntcn} employs DWConv combined with convolutional feed-forward blocks, balancing expressivity and efficiency.

\paragraph{Semiseparable-based}
Semiseparable-based models have emerged as a powerful paradigm for sequence modeling. 
Foundational work, such as S4~\cite{gu2021efficiently} demonstrated efficient state-space layers, 
and subsequent models, including Mamba, SSD~\cite{dao2024transformers} extend this approach with semiseparable factors and content-aware selection mechanisms, S-Mamba~\cite{wang2025mamba}, and CMamba~\cite{zeng2024cmamba} 
achieving strong performance in time series domains.

\paragraph{Motivation for JustDense}    
Various sequence mixers have become the \textit{de facto} standard for time series sequence mixing. 
Nevertheless, recent empirical studies challenge this paradigm. 
DLinear demonstrates that simple linear layers can outperform sophisticated Attention-based models across multiple forecasting benchmarks, 
while TSMixer~\cite{chen2023tsmixer} shows that MLP-based architectures achieve competitive results replacing attention mechanisms with dense layers. 
Similarly, PatchMixer replaces attention with convolutions in PatchTST, achieving improved performance with reduced computational overhead.

However, these studies adopt an \textit{ad hoc} replacement strategy—substituting complex components with simpler alternatives without systematic theoretical justification. 
Current approaches lack a rigorous analysis of the mathematical properties and do not offer principled criteria for determining when complex sequence mixers provide genuine advantages over simpler alternatives.

\section{Preliminaries}
\label{sec: preliminaries}

\begin{figure}[ht!]
\centering 
\includegraphics[width=0.8\columnwidth]{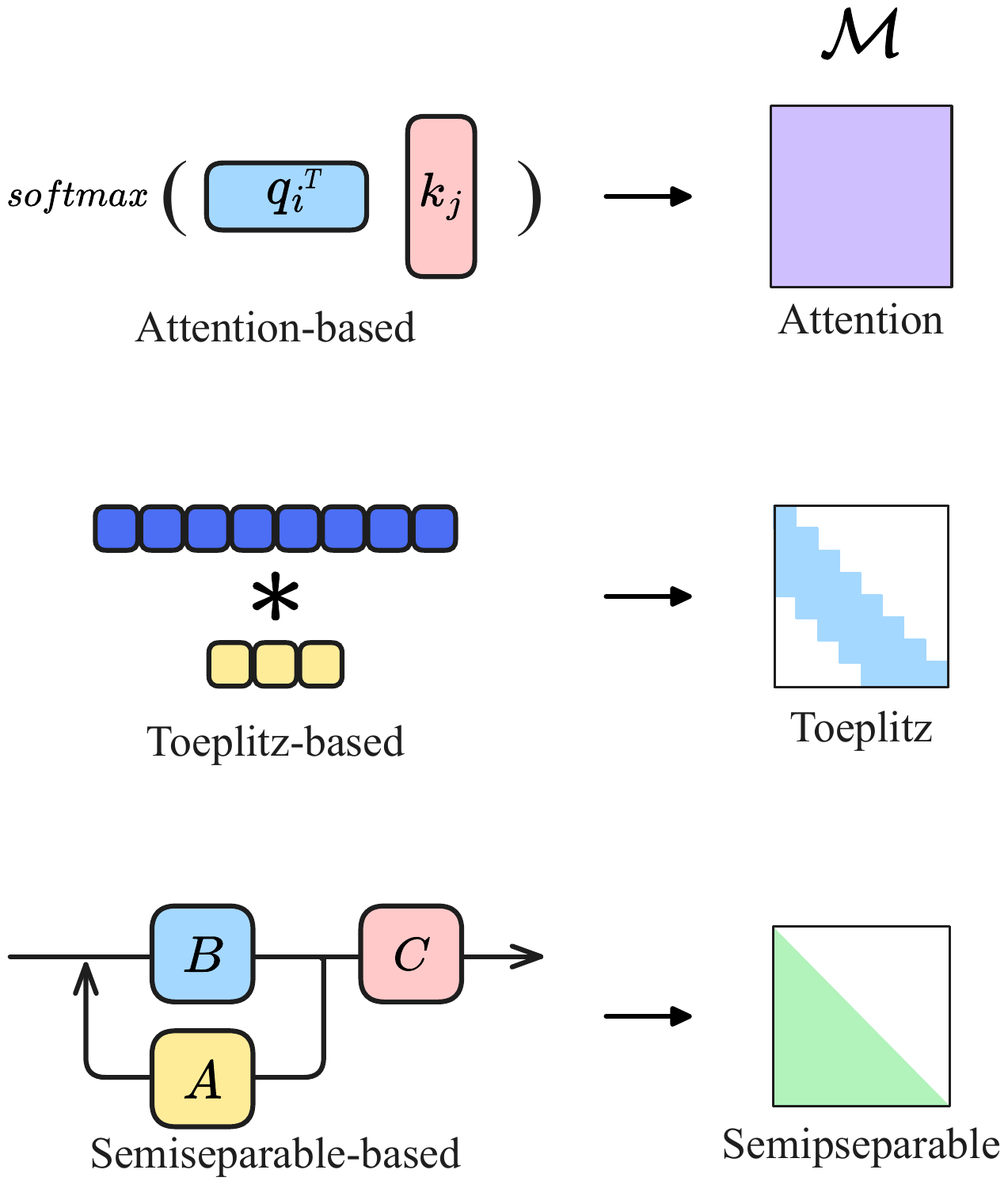}
\caption{Illustration of the class of mixer matrices. The common sequence mixers in TSA models (left) 
can be categorized into three matrix classes $\mathcal{M}$ (right).}
\label{fig:mm}
\end{figure}

\subsubsection{Input Sequence}

Consider a multivariate time series dataset in which the historical observations 
are represented as $\mathbf{X} \in \mathbb{R}^{L \times C}$, 
where $L$ denotes the length of the lookback window and $C$ represents the number of features (or variates). 

\subsubsection{MatrixMixer}
Let \(\mathbf{X}\in\mathbb{R}^{L\times C}\) be the input sequence. The MatrixMixer architecture consists of the following components:
\begin{itemize}
  \item \(f_X\colon\mathbb{R}^{L\times C}\to\mathbb{R}^{L\times D}\): input embedding function,
  \item \(H\): number of heads, \(P\): head dimension, where \(HP=D\),
  \item \(\mathcal{M}\subseteq\mathbb{R}^{L\times L}\): the class of mixer matrices,
  \item  \(f_\mathcal{M}^{(h)}\colon\mathbb{R}^{L\times C}\to\mathcal{M}\): matrix construction function that dynamically generates mixer matrices from input data, parameterized by \(\theta\).
  \item The \textit{MatrixMixer} is defined as: 
  \(\mathbf{M}= f_{\mathcal{M}}(\mathbf{X}, \theta)\;\in\;\mathcal{M}\)
\end{itemize}

For each head \(h=1,\dots,H\), the corresponding head output is computed as
\[
  \mathbf{Y}^{(h)} = \mathbf{M}^{(h)}\,(f_X(\mathbf{X}))^{(h)}
\]

The MatrixMixer framework provides a unified understanding of diverse sequence mixing mechanisms—including attention, semiseparable, 
and toeplitz—by representing them as \(L\times L\) matrices with distinct structural patterns.
Figure \ref{fig:mm} illustrates the concept of common sequence mixers and its corresponding class of mixer matrices \(\mathcal{M}\).

\begin{figure*}[ht!]
\centering
\includegraphics[width=2\columnwidth]{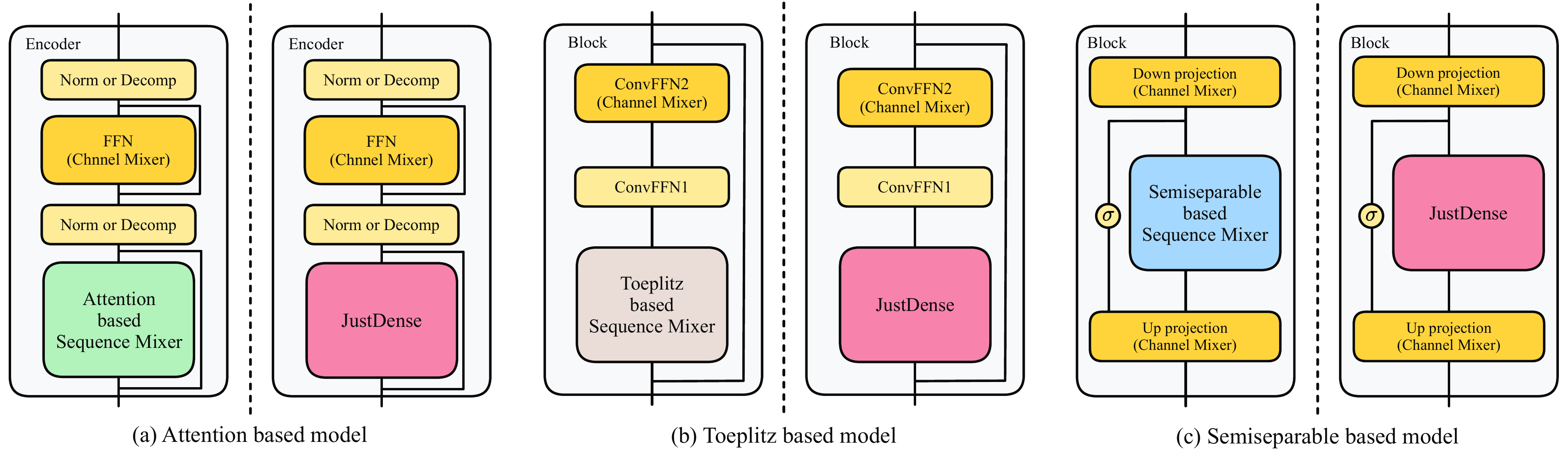}
\caption{Overview of JustDense study design. Based on the \textit{MatrixMixer} framework,
the sequence mixer of attention-based, Toeplitz-based, and semiseparable-based models can be defined as $L\times L$ MatrixMixers.
Subsequently, these mixers are replaced with dense matrices, which are $L\times L$ MatrixMixers.}
\label{fig:overview}
\end{figure*}

\section{JustDense}

In this section, we systematically reformulate sequence mixers as \textit{MatrixMixers} and replace them with equivalent $L\times L$ dense matrices.
In JustDense, the complex mixing operations in attention, toeplitz, and semiseparable-based mixers in TSA models are effectively replaced 
with simple trainable dense transformations without comprising modeling capability.
Figure~\ref{fig:overview} presents an overview of the JustDense, illustrating how structurally common sequence mixers can be unified under dense class mixer matrices.
%
%
Detailed derivations and implementation notes for each model are provided in Appendix~\ref{sec:appendix:methodology}.

\subsection{Attention to Dense}
\label{sec:att2dense}
We interpret the attention of each head, which operates independently on the input sequence, as a \textit{MatrixMixer} (see Figure~\ref{fig:overview}(a)).
Notably, models using attention-based sequence mixer, such as PatchTST and Transformer, share a common architectural pattern: 
(1) an attention-based sequence mixer, (2) a feed-forward network serves as a channel mixer, and (3) normalization or decomposition modules.
Specifically, let \(\mathbf{X}\in\mathbb{R}^{L\times C}\) be the input sequence of length \(L\) with \(C\) channels,
the input sequence is first linearly transformed by:
\begin{equation*}
f_X(\mathbf{X})= \mathbf{X}W_V,
\quad
W_V\in\mathbb{R}^{C\times D}
\end{equation*}
where \(D = H \times P\) is the embedding dimension, such that \(H\) is the number of heads and \(P\) is the per-head dimension.
Here, \(W_V\) is a learnable parameter.
Subsequently, for each head \(h=1,\dots,H\), the query vectors and key vectors are computed as:
\begin{align*}
  Q^{(h)}=\mathbf{X}\,W_Q^{(h)}, \\
  K^{(h)}=\mathbf{X}\,W_K^{(h)}, \\
  \quad W_Q^{(h)},W_K^{(h)}\in\mathbb{R}^{L\times P}
\end{align*}
where $W_Q^{(h)}$ and $W_K^{(h)}$ are learnable parameters.
By employing query and key vectors, the attention score matrix can be defined as $L\times L$ \textit{MatrixMixer} by:
\begin{equation*}
  \mathbf{M}^{(h)}= \mathrm{softmax}\!\Bigl(\tfrac{Q^{(h)}\,{K^{(h)}}^\top}{\sqrt{P}}\Bigr)\;\in\;\mathbb{R}^{L\times L}
\end{equation*}
where the softmax operator is applied for normalization.
The output of each head $h={1,...,H}$ is expressed as
\begin{equation*}
  \mathbf{Y}^{(h)}= \mathbf{M}^{(h)}\,\bigl(f_X(\mathbf{X})\bigr)^{(h)}\;\in\;\mathbb{R}^{L\times P}
\end{equation*}
Notably, \textit{MatrixMixer} shows the dense (softmax attention) class mixer matrices.
\textit{JustDense} replaces each matrix \(\mathbf{M}^{(h)}\) with a learnable dense matrix \(\tilde{\mathbf{M}}^{(h)}\in\mathbb{R}^{L\times L}\) by:
\begin{equation*}
  \mathbf{Y}^{(h)} = \tilde{\mathbf{M}}^{(h)} (f_X(\mathbf{X})^{(h)}) \in \mathbb{R}^{L\times P}
\end{equation*}
Here, each attention head is treated as a dense transformation of the embedded sequence.
Thus, all \(H\) attention heads are replaced with independent \(L\times L\) dense layers, eliminating the need for explicit attention computation.

\subsection{Toeplitz to Dense}
\label{sec:cnn2dense}

ModernTCN~\cite{luo2024moderntcn} employs an encoder with (1) a depth-wise convolutional sequence mixer (DWConv) and (2) a convolutional feed-forward network for channel mixing (ConvFFN1 \& ConvFFN2) (see Figure~\ref{fig:overview}(b)).
Autoformer~\cite{autoformer} employs an encoder with (1) an autocorrelation-based sequence mixer that leverage fourier transformation and convolutions for input decomposition and (2) a feedforward network for channel mixing.
For clarity, this section provides explains the Toeplitz-based sequence mixer in ModernTCN, which can be interpreted as a \textit{MatrixMixer} acting independently for each head.
Let \(\mathbf{X}\in\mathbb{R}^{L \times C}\) denote the input sequence of length \(L\) with \(C\) variables.
Each time series is embedded using non-overlapping patches and downsampling layers to preserve variable independence by:
\begin{align*}
f_X(\mathbf{X}) &= \text{DownsampleLayers}(\mathbf{X}) \\
&= \text{BN}(\text{Conv1d}(\cdots \text{BN}(\text{Conv1d}(\mathbf{X^\top})), \cdots)) \in \mathbb{R}^{B \times D \times L'}
\end{align*}
where Conv1d is a 1D convolutional layer, BN operator is a batch normalization.
The Toeplitz(DWConv) mixer matrix \(\mathbf{M}^{(h)}\in\mathbb{R}^{L'\times L'}\) has entries
\begin{equation*}
  m_{ij}^{(h)} =
  \begin{cases}
    w_{k}^{(h)}, & (i - j)\bmod d = 0,\;k=\tfrac{i - j}{d}+1\in[1,K],\\
    0,           & \text{otherwise},
  \end{cases}
\end{equation*}
where \(K\) is the kernel size, \(d\) the dilation, and \(w_{1:K}^{(h)}\) the per-head kernel weights; thus, it can be 
defined as a \textit{MatrixMixer}.
Notably, the mixer matrix has a single head($H=1$) with a structured parameterization. 
The head output is expressed as
\begin{equation*}
  \mathbf{Y}^{(h)}
  = \mathbf{M}^{(h)}\,(f_X(\mathbf{X}))^{(h)}
  \;\in\;\mathbb{R}^{L'\times P},
\end{equation*}

According to the MatrixMixer framework, \textit{MatrixMixer} can be categorized as a Toeplitz class mixer matrices.
To generalize this mixer, \textit{JustDense} replaces each structured \(\mathbf{M}^{(h)}\) with a fully learnable dense matrix \(\tilde{\mathbf{M}}^{(h)}\in\mathbb{R}^{L'\times L'}\), yielding
\begin{equation*}
  \mathbf{Y}^{(h)}
  = \tilde{\mathbf{M}}^{(h)}\,(f_X(\mathbf{X}))^{(h)}
  \;\in\;\mathbb{R}^{L'\times P},
\end{equation*}
such that each head performs a simple dense transformation of the embedded sequence, eliminating convolution-specific structure while preserving variable independence.

\subsection{Semiseparable to Dense}
\label{sec:ssm2dense}
Models, such as Mamba1, SSD, and S-Mamba, use (1) a state-space model (SSM) for sequence mixing and (2) a pre-projection channel mixer~\cite{beck:24xlstm} (Figure~\ref{fig:overview}(c)).
For clarity, we focus on the SSD, which employs semiseparable-based sequence mixer.
Specifically, the input sequence \(\mathbf{X}\in\mathbb{R}^{L\times C}\) is first linearly transformed by:
\begin{equation*}
f_X(\mathbf{X}) = \mathbf{X}W_{\mathrm{emb}},
\quad
W_{\mathrm{emb}}\in\mathbb{R}^{C\times D}
\end{equation*}
where \(D\) is the embedding dimension.
For each head \(h=1,\dots,H\), the Semiseparable mixer matrix \(\mathbf{M}^{(h)}\in\mathbb{R}^{L\times L}\) has a structured parameterization based on state-space dynamics, where each \((i,j)\)-element \(m_{ij}^{(h)}\) is defined as:
\begin{equation*}
m_{ij}^{(h)} = c_i^{(h)\top} \prod_{k=j+1}^{i-1} A_k^{(h)} b_j^{(h)}
\end{equation*}
where \(A_k^{(h)} \in \mathbb{R}^{P \times P}\) are state transition matrices, and \(c_i^{(h)}, b_j^{(h)} \in \mathbb{R}^{P}\) are output and input vectors respectively, computed as:
\begin{align}
c_i^{(h)} &= \mathbf{x}_i W_C^{(h)}, \quad W_C^{(h)} \in \mathbb{R}^{C \times P} \\
b_j^{(h)} &= \mathbf{x}_j W_B^{(h)}, \quad W_B^{(h)} \in \mathbb{R}^{C \times P}
\end{align}
This formulation captures the sequential dependencies through a learnable state-space representation. 
The head output is then
\begin{equation*}
  \mathbf{Y}^{(h)} = \mathbf{M}^{(h)}\,(f_X(\mathbf{X}))^{(h)} \;\in\;\mathbb{R}^{L\times P}
\end{equation*}

According to the MatrixMixer framework, \textit{MatrixMixer} is categorized as a Semiseparable class mixer matrices.
To generalize this mixer, \textit{JustDense} replaces each structured semiseparable matrix \(\mathbf{M}^{(h)}\) with a fully learnable dense matrix \(\tilde{\mathbf{M}}^{(h)}\in\mathbb{R}^{L\times L}\), assuming
\begin{equation*}
  \mathbf{Y}_{\mathrm{dense}}^{(h)} = \tilde{\mathbf{M}}^{(h)}\,(f_X(\mathbf{X}))^{(h)} \;\in\;\mathbb{R}^{L\times P}
\end{equation*}
such that each head performs a simple dense transformation of the embedded sequence, eliminating the complex state-space dynamics while preserving the essential sequence modeling capabilities.

\begin{table*}[ht!]
\centering
\setlength\tabcolsep{3.5pt}
\footnotesize
\begin{tabular}{llcccccccccccccc}
\toprule
\multirow{3}{1.5cm}{\textbf{Tasks}} & \multirow{3}{*}{\textbf{Metrics}} & \multicolumn{2}{c}{\textbf{Transformer}} & \multicolumn{2}{c}{\textbf{Autoformer}} & \multicolumn{2}{c}{\textbf{PatchTST}} & \multicolumn{2}{c}{\textbf{Mamba}} & \multicolumn{2}{c}{\textbf{iTransformer}} & \multicolumn{2}{c}{\textbf{S-Mamba}} & \multicolumn{2}{c}{\textbf{ModernTCN}} \\[0.1em]
& & \multicolumn{2}{c}{\citeyear{vaswani2017transformer}} & \multicolumn{2}{c}{\citeyear{autoformer}} & \multicolumn{2}{c}{\citeyear{Yuqietal-2023-PatchTST}} & \multicolumn{2}{c}{\citeyear{gu2023mamba}} & \multicolumn{2}{c}{\citeyear{liu2024itransformerinvertedtransformerseffective}} & \multicolumn{2}{c}{\citeyear{wang2025mamba}} & \multicolumn{2}{c}{\citeyear{luo2024moderntcn}} \\[0.1em]
\cmidrule(lr){3-4} \cmidrule(lr){5-6} \cmidrule(lr){7-8} \cmidrule(lr){9-10} \cmidrule(lr){11-12} \cmidrule(lr){13-14} \cmidrule(lr){15-16}
& & Orig. & JD & Orig. & JD & Orig. & JD & Orig. & JD & Orig. & JD & Orig. & JD & Orig. & JD \\[0.2em]
\midrule
Classification & Acc. & 0.721 & \textbf{0.723} & 0.613 & \textbf{0.672} & 0.867 & \textbf{0.691} & 0.693 & \textbf{0.716} & \textbf{0.710} & 0.706 & 0.717 & \textbf{0.721} & 0.692 & \textbf{0.703} \\[1.0em]
\makecell[l]{Anomaly Detection} & F1 & 0.783 & \textbf{0.791} & 0.779 & \textbf{0.794} & 0.828 & \textbf{0.843} & 0.774 & \textbf{0.781} & 0.817 & \textbf{0.824} & \textbf{0.822} & 0.818 & 0.836 & \textbf{0.846} \\[1.0em]
Imputation & MSE & 0.171 & \textbf{0.144} & 0.390 & \textbf{0.171} & 0.068 & \textbf{0.066} & 0.164 & \textbf{0.108} & 0.103 & \textbf{0.101} & 0.114 & \textbf{0.105} & \textbf{0.04} & \textbf{0.04} \\[1.0em]
\makecell[l]{Long-Term} & MSE & 1.133 & \textbf{1.067} & 0.457 & \textbf{0.445} & \textbf{0.348} & 0.358 & 0.358 & \textbf{0.348} & 0.323 & \textbf{0.292} & 0.370 & \textbf{0.348} & \textbf{0.342} & \textbf{0.342} \\[1.0em]
\makecell[l]{Short-Term} & MASE & \textbf{16.817} & 16.822 & 2.459 & \textbf{2.139} & 1.740 & \textbf{1.654} & 1.629 & \textbf{1.616} & 1.757 & \textbf{1.625} & \textbf{1.605} & 1.618 & \textbf{1.605} & 1.618 \\[0.5em]
\bottomrule
\end{tabular}
\vspace{0.5em}
\caption{Performance comparison across five core time-series tasks. Boldfaced values indicate the higher performance between the original (Orig.) and JustDense applied (JD)
models.}
\label{tab:main_results}
\end{table*}

\section{Experiments}
In this section, we evaluate seven state-of-the-art TSA models across five TSA tasks by converting their sequence mixers into dense matrix mixers, and assessing the impact on performance. 
Furthermore, we compared the efficiency and analyzed the structural properties of the learned dense mixers against the original sequence mixers.

\subsubsection{Datasets}
For experiments, we selected a diverse set of time series datasets spanning  various domains and tasks:
\textbf{Classification} (10 datasets) covering gesture recognition, health monitoring, and voice analysis~\cite{bagnall2018uea};
\textbf{Anomaly Detection} (5 datasets) from industrial monitoring systems~\cite{su2019robust, hundman2018detecting, mathur2016swat, abdulaal2021practical};
\textbf{Imputation} (4 datasets) focusing on electricity and environmental data~\cite{haoyietal2021informer, electricityloaddiagrams20112014_321, autoformer};
\textbf{Long-term Forecasting} (7 datasets), including electricity, weather, traffic, and economic time series~\cite{haoyietal2021informer, electricityloaddiagrams20112014_321, autoformer,lai2018modeling}; and
\textbf{Short-term Forecasting} (6 datasets) from the M4 competition spanning multiple frequencies~\cite{makridakis2018m4}.
These datasets exhibit significantly variation in dimensionality (1-963 features),
sequence length (6-69,680 time steps), and sampling frequencies, providing comprehensive coverage of real-world time series analysis scenarios.
Detailed dataset specifications, including dimensions, lengths, and domain descriptions, are provided in Appendix~\ref{appendix:datasets}.

\subsubsection{Implementation} 
All experiments were implemented using PyTorch 2.7.1+cu126~\cite{paszke2019pytorch} and conducted on a single NVIDIA L40S D6 48GB GPU with an Intel(R) Xeon(R) Gold 6426Y processor.
Code is available at: \url{https://github.com/Thrillcrazyer/JustDense}

\subsubsection{Baselines}
For comparison, we selected seven representative TSA models whose sequence mixer can be defined as MatrixMixer.
These models include Transformer~\cite{vaswani2017transformer}, Autoformer~\cite{autoformer}, ModernTCN~\cite{luo2024moderntcn}, 
PatchTST~\cite{gong2023patchmixer}, Mamba~\cite{gu2023mamba}, S-Mamba~\cite{wang2025mamba}, and iTransformer~\cite{liu2024itransformerinvertedtransformerseffective}.

\subsection{Performance Comparison}
\label{exp:performance_comparison}

Table~\ref{tab:main_results} demonstrates that converting original sequence mixers into dense matrix consistently improves performance across all five core tasks. 
The performance is averaged across each task for each model. 
For most models and tasks, the JustDense (JD) variant outperforms the original (Orig.) version, as indicated by the bold values. 
This includes higher accuracy for classification, higher F1 scores for anomaly detection, and lower error metrics (MSE, MASE) for imputation and forecasting tasks.
Note that employing dense matrix instead of sequence mixers can actually enhance various TSA models regardless of their underlying mechanisms.
There are various reasons for this observation.
First, dense matrix operates in the full\(L\times L\) matrix space—superseding low-rank attention
(\(\operatorname{rank}(attention)\le P\)), toeplitz (\(\operatorname{rank}(toeplitz)\le K+1\)),
and semiseparable (\(\operatorname{rank}(semiseparable)\le N\lceil L/N\rceil\)) mixers—to capture rich, high-rank interactions.
Moreover, implicit bias of gradient descent toward minimal nuclear-norm solutions ensures that dense matrix naturally recovers structured,
low-rank behavior when appropriate while modeling additional dependencies.
(For tight rank bounds and the formal low-displacement-rank inclusion argument, see Appendix~\ref{appendix:rank}.)

\subsection{Efficiency Comparison}
\label{exp:efficiency_comparison}

\begin{figure}[ht!]
\centering
\includegraphics[width=1\columnwidth]{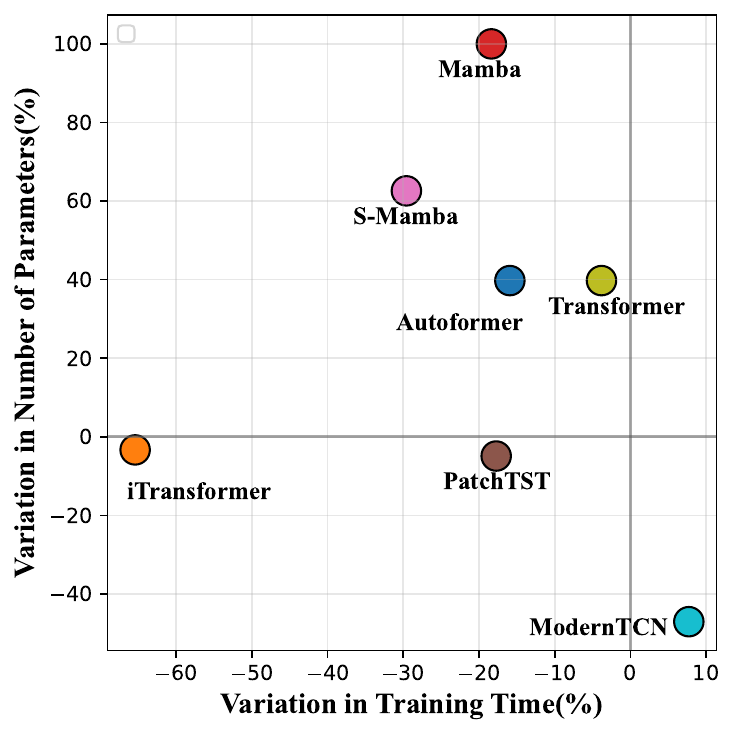}
\caption{Efficiency comparison in terms of training time (x-axis) and the number of parameters (y-axis). 
The results are from the long-term forecasting task on the ETTm2 dataset.}
\label{fig:speed}
\end{figure}

We compare the memory usage and training speed of time series models before and after converting the sequence mixers to dense matrix mixers
on the long-term forecasting task using the ETTm2 dataset. The results are summarized in Figure~\ref{fig:speed}.

Our analysis reveals that for Encoder-Decoder structured models, such as Autoformer and Transformer, the number of parameters increases while training time decreases. 
This is because replacing the self-attention mechanism with a dense layer increases parameter count, but eliminates the computational overhead of the attention mechanism, thereby reducing training time.

For Encoder-only models, such as iTransformer and PatchTST, converting to a dense layer results in a reduction in both parameter count and training time. The dense layer requires fewer parameters than those needed for the Q and K computations in self-attention, and the architecture is simplified by removing the attention mechanism.

For SSM-based models, such as Mamba and S-Mamba, converting to dense layers leads to an increase in parameters but decreased training time. This is because the number of parameters required for a dense layer to act as a sequence mixer is theoretically larger. Although SSM-based models are theoretically more computationally efficient, the dense operations are faster in practice. This is probably owing to the use of highly optimized libraries, such as cuBLAS~\cite{nvidia_cublas_129}, which perform parallel computations more simply and efficiently.

For the convolution-based model, ModernTCN, the training time increased while the number of parameters decreased. In ModernTCN, DWConvolution performs sequence mixing on a large tensor $\mathbf{T} \in C \times P \times H$. When converted to a dense operation, the parameter count is reduced. This outcome aligns with the theoretical understanding that convolution operations are generally faster than dense operations.

\begin{figure}[ht!]
\begin{subfigure}[b]{0.45\textwidth}
    \centering
    \includegraphics[width=\textwidth]{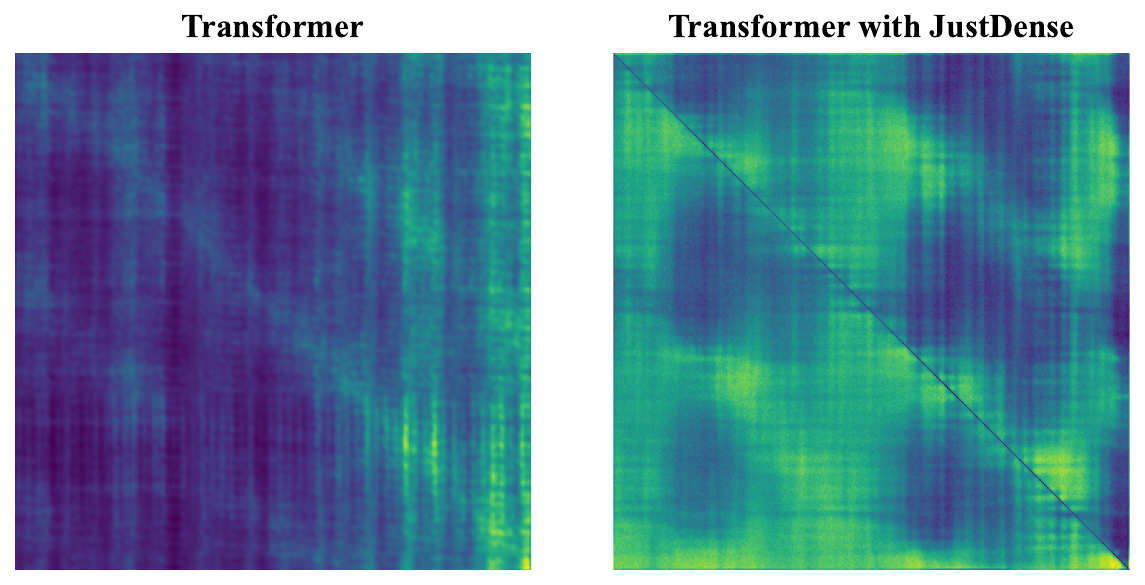}
    \caption{Transformer: Attention Matrix vs. Dense Matrix}
    \label{fig:attention_vs_dense_Transformer}
\end{subfigure}

\begin{subfigure}[b]{0.45\textwidth}
    \centering
    \includegraphics[width=\textwidth]{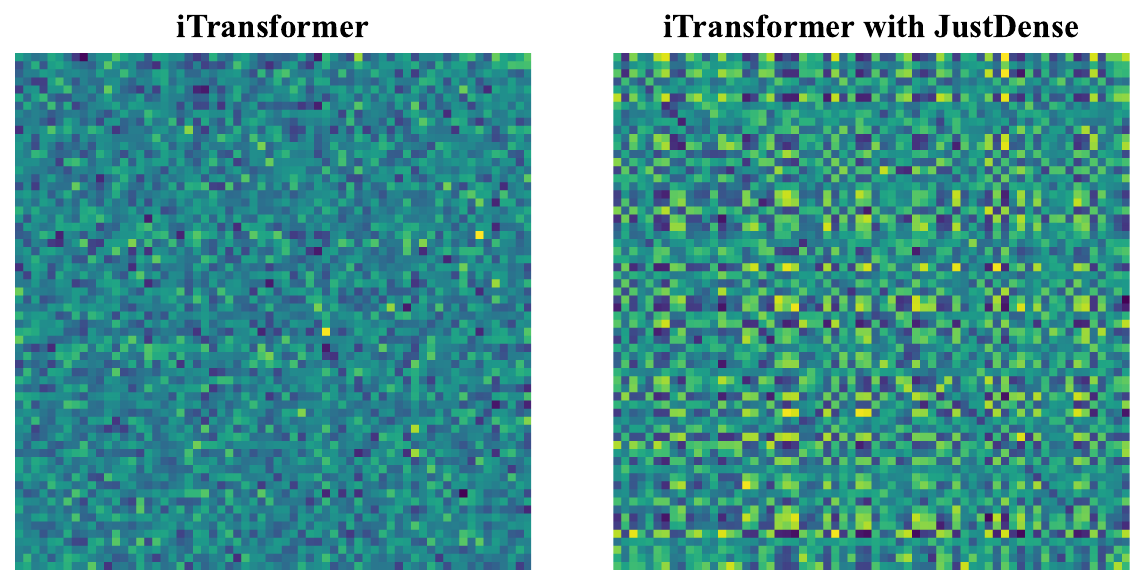}
    \caption{iTransformer: Attention Matrix vs. Dense Matrix}
    \label{fig:attention_vs_dense_iTransformer}
\end{subfigure}

\begin{subfigure}[b]{0.45\textwidth}
    \centering
    \includegraphics[width=\textwidth]{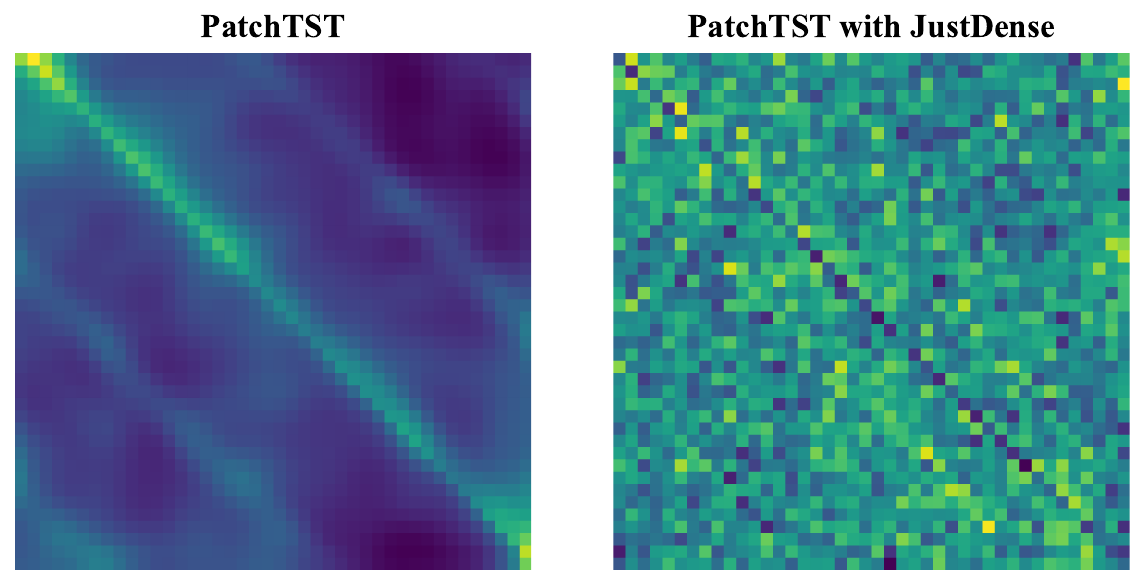}
    \caption{PatchTST: Attention Matrix vs. Dense Matrix}
    \label{fig:attention_vs_dense_PatchTST}
\end{subfigure}

\begin{subfigure}[b]{0.45\textwidth}
    \centering
    \includegraphics[width=\textwidth]{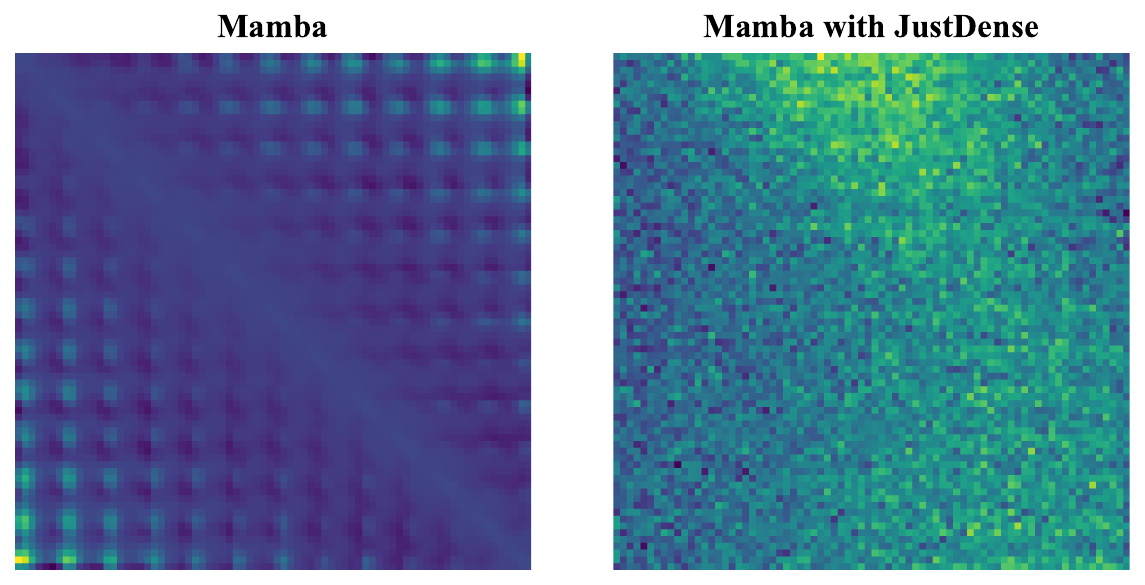}
    \caption{Mamba: SSM Matrix vs. Dense Matrix}
    \label{fig:attention_vs_dense_Mamba}
\end{subfigure}

\begin{subfigure}[b]{0.45\textwidth}
    \centering
    \includegraphics[width=\textwidth]{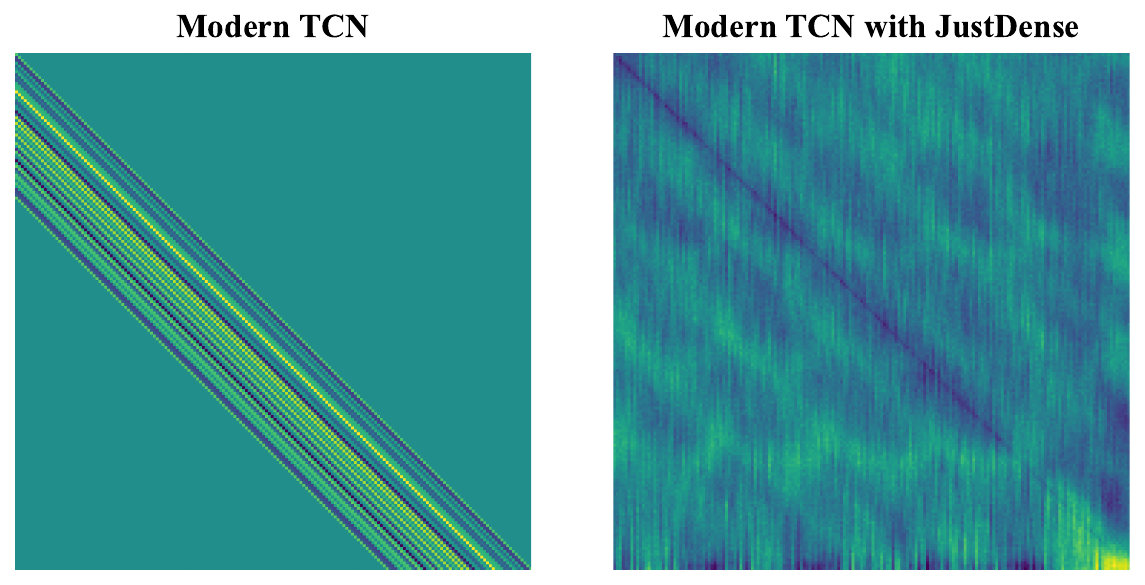}
    \caption{ModernTCN: Toeplitz Matrix vs. Dense Matrix}
    \label{fig:attention_vs_dense_ModernTCN}
\end{subfigure}
\caption{Visualization of similarity between the original matrix mixer (right) and converted dense matrix (left).}
\label{fig:formal_vs_dense}
\end{figure}

\subsection{Attention vs. Dense Matrix Comparison}
\label{exp:attention_vs_dense}

\begin{table}[ht!]
\centering
\setlength\tabcolsep{7pt}
\begin{tabular}{lcc}
\toprule
\textbf{Model} & \multicolumn{2}{c}{\textbf{Metric}} \\
\cmidrule(lr){2-3}
 & PSNR & JSD \\
\midrule
Transformer (\citeyear{vaswani2017transformer}) & 16.007 & 0.06 \\
PatchTST (\citeyear{Yuqietal-2023-PatchTST}) & 14.165 & 0.05 \\
Mamba (\citeyear{gu2023mamba}) & 4.18 & 0.06 \\
iTransformer (\citeyear{liu2024itransformerinvertedtransformerseffective}) & 8.269 & 0.03 \\
ModernTCN (\citeyear{luo2024moderntcn}) & 20.665 & 0.01 \\
\bottomrule
\end{tabular}
\caption{Similarity between original matrix mixer and converted dense matrix.}
\label{exp:attention_vs_dense_comparison}
\end{table}

We compared the original MatrixMixer and the Dense MatrixMixer converted by JustDense.
For this comparison, we employed the long-term forecasting task on the Weather dataset.
First, we trained the original models and dense-converted models.
Subsequently, we extracted the sequence mixers from the trained models and visualize and compared them.
We used Peak signal-to-noise ratio (PSNR) and Jensen-Shannon divergence (JSD) as metrics. 
PSNR measures visual distribution differences and JSD quantifies differences in probability distributions.
As two matrices are similar, PSNR increases and JSD decreases.

The results of comparison are shown in Table~\ref{exp:attention_vs_dense_comparison}.
The results show that the PSNR values are relatively low, while the JSD values are significantly small.
Thus, the original MatrixMixer and the Dense MatrixMixer converted by JustDense are visually different but maintain the essential information of the original sequence.
Therefore, even after conversion, the model effectively preserves the key patterns of the original sequence.
Specifically, despite structural changes, the essential information is well maintained.

Figure~\ref{fig:formal_vs_dense} shows the formal matrix and the dense matrix converted by JustDense.
This visualization shows that the dense matrix captures temporal relationships in a manner similar to the original formal matrix mixer.

\section{Conclusion}
\label{sec:conclusion}
In this study, we systematically question the necessity of complex sequence mixers in time series analysis. 
We reformulated various sequence mixers as simple $L \times L$ dense layers—termed MatrixMixers—and conducted an extensive empirical study, named \textit{JustDense}, 
across diverse architectures and benchmark datasets. 
Our evaluation includes performance, efficiency, and structural analyses comparing the original models and their dense matrix counterparts.
Through the experiments, we provided strong empirical evidence that the complexity of modern sequence mixers is unnecessary for time series data.
Our results reveal that converting complex sequence mixers into dense matrix, despite their simplicity, can match or even surpass the performance.
Dense matrix, as high-rank matrices, can effectively capture both global and local dependencies without overfitting 
owing to their regularization effects and other architectural components.
Structural analyses further confirm that dense layers perform similar representational roles to specialized mixers.
In terms of efficiency, although dense layers have theoretical $O(n^2)$ complexity, 
practical implementation benefits from parallel computation and the elimination of overhead associated with dynamically constructing sparse or structured matrices. 
Consequently, replacing sequence mixers with dense matrix often yields competitive or superior efficiency in real-world settings.
However, while dense matrix demonstrate strong empirical performance, 
their large number of parameters may require other mixermatrices structure to fit the data. Exploring more effective other architectures
or data fitting Matrixmixers remains a promising direction for future research.

Despite these considerations, our findings still provide compelling evidence against the prevailing assumption that 
“\textbf{deeper and more complex is better}” in time series modeling. 
This suggests that the unique characteristics of time series data may require fundamentally different design principles 
than those derived from natural language processing fields.


\bibliography{aaai2026}

\newpage
\clearpage
\appendix

\section*{Appendix}

\section{Detailed Description of Datasets}
\label{appendix:datasets}
\begin{table*}[t] 
  \centering
  \setlength\tabcolsep{7pt}
  \begin{tabular}{llllll}
    \toprule
    \textbf{Task}                           & \textbf{Name}           & \textbf{Dimension}  & \textbf{Length}   & \textbf{Domain} \\
    \midrule
    \multirow{10}{*}{Classification}        & EthanolConcentration    & 3                   & 1,751             & Alcohol Industry \\
                        & FaceDetection           & 144                 & 62                & Face (250 Hz) \\
                        & Handwriting             & 3                   & 152               & Motion \\
                        & Heartbeat               & 61                  & 405               & Health (0.061 secs) \\
                        & JapaneseVowels          & 12                  & 29                & Voice \\
                        & PEMS-SF                 & 963                 & 144               & Transportation (1 day) \\
                        & SelfRegulationSCP1      & 6                   & 896               & Health (256 Hz) \\
                        & SelfRegulationSCP2      & 7                   & 1,152             & Health (256 Hz) \\
                        & SpokenArabicDigits      & 13                  & 93                & Voice (11025 Hz) \\
                        & UWaveGestureLibrary     & 3                   & 315               & Gesture \\
    \midrule
    \multirow{5}{*}{Anomaly Detection}      & SMD                     & 38                  & 100               & Industry (1 min) \\
                        & MSL                     & 55                  & 100               & Industry (1 min) \\
                        & SMAP                    & 25                  & 100               & Industry (1 min) \\
                        & SWaT                    & 51                  & 100               & Industry (1 min) \\
                        & PSM                     & 25                  & 100               & Industry (1 min) \\
    \midrule
    \multirow{4}{*}{Imputation}             & ETTh1, ETTh2            & 7                   & 17,420            & Electricity (1 hour) \\
                        & ETTm1, ETTm2            & 7                   & 69,680            & Electricity (15 mins) \\
                        & Electricity             & 321                 & 26,304            & Electricity (1 hour) \\
                        & Weather                 & 21                  & 52,696            & Environment (10 mins) \\
    \midrule
    \multirow{5}{*}{Long-term Forecasting}  & ETTh1, ETTh2            & 7                   & 17,420            & Electricity (1 hour) \\
                        & ETTm1, ETTm2            & 7                   & 69,680            & Electricity (15 mins) \\
                        & Electricity             & 321                 & 26,304            & Electricity (1 hour) \\
                        & Weather                 & 21                  & 52,696            & Environment (10 mins) \\
                        & Traffic                 & 862                 & 17,544            & Transportation (1 hour) \\
                      
    \midrule
    \multirow{6}{*}{Short-term Forecasting} & M4-Yearly               & 1                   & 6                 & Demographic \\
                        & M4-Quarterly            & 1                   & 8                 & Finance \\
                        & M4-Monthly              & 1                   & 18                & Industry \\
                        & M4-Weakly               & 1                   & 13                & Macro \\
                        & M4-Daily                & 1                   & 14                & Micro \\
                        & M4-Hourly               & 1                   & 48                & Other \\
    \bottomrule
  \end{tabular}
  \caption{Summary of Time Series Datasets Used in This Paper}
  \label{tb:datasets}
\end{table*}

This section outlines the data characteristics and selection criteria for the datasets used in the various time series tasks addressed in this study.
As discussed in the main body, benchmarking across tasks, including classification, imputation, forecasting, and anomaly detection, present unique challenges.
To ensure a comprehensive evaluation, we select widely-recognized real-world datasets covering diverse domains.
Specifically, the datasets used in this study follow the benchmark protocol established in~\cite{wang2024deep}.

We provide the detailed metadata for these datasets in Table~\ref{tb:datasets}.

For the \textbf{classification tasks},
we utilized the popular UEA archive, comprising datasets from diverse domains, including health, voice recognition, gestures, and transportation~\cite{bagnall2018uea}.
These datasets vary significantly in dimensionality, ranging from low-dimensional gesture data to high-dimensional sensor data.
Each dataset has clearly defined training and test splits as outlined by the original benchmark.

For the \textbf{anomaly detection tasks}, we selected industrially relevant datasets, such as
\textit{Server Machine Dataset (SMD)}, \textit{Mars Science Laboratory rover (MSL)},
\textit{Soil Moisture Active Passive satellite (SMAP)}, \textit{Secure Water Treatment (SWaT)},
and \textit{Pooled Server Metrics (PSM)}~\cite{su2019robust, hundman2018detecting, mathur2016swat, abdulaal2021practical}.
These datasets comprise continuous multivariate time series collected at fine-grained resolutions (typically 1-minute intervals),
with sliding window sampling employed for anomaly detection benchmarking.

For the \textbf{imputation tasks}, we selected datasets commonly used for evaluating time series imputation methods,
specifically, the \textit{Electricity Transformer Temperature (ETT)}, \textit{Electricity},
and \textit{Weather} datasets~\cite{haoyietal2021informer, electricityloaddiagrams20112014_321, autoformer}.
ETT includes four sub-datasets (\textit{ETTh1}, \textit{ETTh2}, \textit{ETTm1}, and \textit{ETTm2}),
collected from electricity transformers at two different resolutions (hourly and 15-minute intervals).
The \textit{Weather} dataset provides comprehensive meteorological observations collected every 10 minutes,
while \textit{Electricity} dataset includes hourly consumption records from multiple clients.

For the \textbf{long-term forecasting tasks}, we leveraged the ETT, \textit{Electricity}, and \textit{Weather} datasets,
considering their extended time horizons and multivariate characteristics~\cite{haoyietal2021informer, electricityloaddiagrams20112014_321, autoformer}.
Additionally, we included three widely recognized datasets: \textit{Traffic}, capturing occupancy rates from 862 freeway sensors at hourly intervals;
\textit{Exchange}, consisting of daily exchange rates for eight countries over multiple decades;
and the influenza-like illness dataset (\textit{ILI}), collected weekly from health agencies~\cite{autoformer, lai2018modeling}.

Finally, for the \textbf{short-term forecasting tasks}, we adopted the M4 benchmark dataset,
a large-scale collection of time series data from different domains,
including demographics, finance, industry, and macroeconomics~\cite{makridakis2018m4}.
In contrast to the long-term datasets that involve continuous sliding window sampling from single-source time series,
M4 comprises numerous independent series with significantly varying temporal properties, making short-term forecasting more challenging.
Each subset of M4 (Yearly, Quarterly, Monthly, Weekly, Daily, and Hourly) has clearly defined prediction lengths, reflecting typical short-term forecasting horizons.

\section{Detailed Formulations: From Sequence Mixers to Dense Representations}
\label{sec:appendix:methodology}
This section provides formal descriptions for the mathematical representation of sequence mixers.
Aligning with Preliminaries, each sequence mixer for head \(h\) is defined by
\begin{equation*}
  \mathbf{M}^{(h)} \;=\; f_{\mathcal{M}}^{(h)}\bigl(\mathbf{X},\theta\bigr)\;\in\;\mathbb{R}^{L\times L}
\end{equation*}
where \(f_{\mathcal{M}}^{(h)}\) is the original mixer-matrix construction function depending on parameters $\theta$. In JustDense, the original mixer-matrix is replaced with
\begin{equation*}
  \tilde{\mathbf{M}}^{(h)} \;=\; f_{\mathrm{Dense}}^{(h)}(\mathbf{X})
  \;\in\;\mathbb{R}^{L\times L}
\end{equation*}
a learnable dense mixer, parameterized by its own learnable weights.

\subsection{Transformer}
Let the historical observations be \(\mathbf{X}\in\mathbb{R}^{L\times C}\), where \(L\) is the lookback length and \(C\) is the number of features.
We first project \(\mathbf{X}\) into a \(D\)-dimensional feature space:
\begin{equation*}
  f_X(\mathbf{X}) = \mathbf{X} W_V + W_{\mathrm{pos}},
  \quad
  W_V \in \mathbb{R}^{C \times D}, \quad
  W_{\mathrm{pos}} \in \mathbb{R}^{L \times D}
\end{equation*}
where \(D=H\,P\), with \(H\) heads of dimension \(P\).

For each head \(h=1,\dots,H\), define the query and key matrices:
\begin{align*}
    Q^{(h)} = f_X(\mathbf{X})\,W_Q^{(h)}, \\
    K^{(h)} = f_X(\mathbf{X})\,W_K^{(h)}, \\
    \quad W_Q^{(h)}, W_K^{(h)} \in \mathbb{R}^{D \times P}
\end{align*}

The sequence mixer matrix for each head is expressed as:
\begin{equation*}
  \mathbf{M}^{(h)} = \mathrm{softmax}\left(\frac{Q^{(h)} {K^{(h)}}^\top}{\sqrt{P}}\right)
  \quad \in \mathbb{R}^{L \times L}
\end{equation*}
This attention-based mixer is parameterized by the learnable weight matrices \(W_Q^{(h)}\) and \(W_K^{(h)}\),
which project the input into the query and key spaces.
The head output is:
\begin{equation*}
  \mathbf{Y}^{(h)} = \mathbf{M}^{(h)} \cdot \left(f_X(\mathbf{X})\right)^{(h)} \in \mathbb{R}^{L \times P}
\end{equation*}

Here, \textit{MatrixMixer} show the Attention class mixer matrices.

In JustDense, each attention-based mixer $\mathbf{M}^{(h)}$ is replaced by a learnable dense matrix
\begin{equation*}
  \tilde{\mathbf{M}}^{(h)} = f_{\mathrm{Dense}}^{(h)}(\mathbf{X}) \in \mathbb{R}^{L\times L}
\end{equation*}
then the head output becomes
\begin{equation*}
  \mathbf{Y}_{\mathrm{dense}}^{(h)} = \tilde{\mathbf{M}}^{(h)} \bigl(f_X(\mathbf{X})\bigr)^{(h)} \in \mathbb{R}^{L\times P}
\end{equation*}

This approach maintains the sequence-level mixing structure (\(\mathbf{M}^{(h)} \in \mathbb{R}^{L \times L}\)) while eliminating the need for query-key attention mechanisms, enabling full replacement of attention-based mixers with simple dense matrices acting along the sequence dimension.

\subsection{PatchTST}
Let the raw sequence be \(\mathbf{X}\in\mathbb{R}^{L\times C}\). We first partition into non-overlapping patches of length \(\ell\), yielding
\begin{equation*}
  L_p = \frac{L}{\ell}, 
  \quad
  \hat{\mathbf{X}} \in \mathbb{R}^{L_p \times (\ell\,C)}
\end{equation*}
where each row of \(\hat{\mathbf{X}}\) concatenates \(\ell\) consecutive time steps across all \(C\) channels. Thereafter, we embed:
\begin{equation*}
  f_X(\hat{\mathbf{X}})= \hat{\mathbf{X}}\,W_V + W_{\mathrm{pos}}', \;
  W_V\in\mathbb{R}^{(\ell\,C)\times D}, \;
  W_{\mathrm{pos}}'\in\mathbb{R}^{L_p\times D}
\end{equation*}
with \(D = H\,P\) and \(P\) the head dimension.  

For each head \(h=1,\dots,H\), compute:
\begin{equation*}
  \begin{aligned}
    Q^{(h)} = f_X(\hat{\mathbf{X}})\,W_Q^{(h)}, \\
    K^{(h)} = f_X(\hat{\mathbf{X}})\,W_K^{(h)}, \\
    W_Q^{(h)},W_K^{(h)}\in\mathbb{R}^{D\times P}
  \end{aligned}
\end{equation*}
The channel-independent self-attention mixer is
\begin{equation*}
\mathbf{M}^{(h)}
= \mathrm{softmax}\!\Bigl(\frac{Q^{(h)}\,{K^{(h)}}^\top}{\sqrt{P}}\Bigr)
\;\in\;\mathbb{R}^{L_p\times L_p}
\end{equation*}
The mixer is fully determined by learnable matrices \(W_Q^{(h)}\) and \(W_K^{(h)}\),
which project each patch embedding into query and key representations.
Each head output is
\begin{equation*}
\mathbf{Y}^{(h)}= \mathbf{M}^{(h)}\,\bigl(f_X(\hat{\mathbf{X}})\bigr)^{(h)}
\;\in\;\mathbb{R}^{L_p\times P}
\end{equation*}

According to MatrixMixer framework, \textit{MatrixMixer} show the softmax Attention class mixer matrices.

In JustDense, each self-attention mixer $\mathbf{M}^{(h)}$ over patches is replaced by
\begin{equation*}
  \tilde{\mathbf{M}}^{(h)} = f_{\mathrm{Dense}}^{(h)}(\hat{\mathbf{X}}) \in \mathbb{R}^{L_p\times L_p}
\end{equation*}
then
\begin{equation*}
  \begin{aligned}
  \mathbf{Y}_{\mathrm{dense}}^{(h)} &= \tilde{\mathbf{M}}^{(h)}\bigl(f_X(\hat{\mathbf{X}})\bigr)^{(h)}
  \end{aligned}
\end{equation*}

\subsection{iTransformer}
Let the historical observations be
\[
\mathbf{X} \in \mathbb{R}^{L\times C}
\]
and transpose into
\[
R = \mathbf{X}^\top \in \mathbb{R}^{C\times L}
\]

Feed \(R\) into a softmax-based attention module to mix information across channels:
\[
Y = \mathrm{Attention}(R) \in \mathbb{R}^{C\times D}, \quad D = H\cdot P
\]
The attended output \(Y\) is added residually to \(R\) and passed through layer normalization.

Next, we treat the FFN that follows attention as the \textit{sequence mixer}. For each position in \(Y\), the mixer matrix
\[
\mathbf{Z} = W_{\mathrm{down}}\;\sigma\bigl(W_{\mathrm{up}}\mathbf{Y}\bigr)
\quad\in\quad\mathbb{R}^{D\times D}
\]

$W_{\mathrm{down}}$, \(W_{\mathrm{up}}\in\mathbb{R}^{D\times U}\) are learnable projection matrices, where \(U\) is the FFN hidden dimension.

\(\sigma(\cdot)\) is the ReLU-like activation function, and \(\sigma\) can be regarded as a type of mask matrix \(D_\text{mask}\).
It is defined as:

\begin{equation*}
  D_{ij} =
  \begin{cases}
    1
    & \begin{aligned}[t]
        \text{if } &  W_{\mathrm{up}}\mathbf{X} > 0 
      \end{aligned} \\
    0, & \text{otherwise}
  \end{cases}
\end{equation*}

\textit{MatrixMixer} is :

\[
\mathbf{M} = W_{\mathrm{down}} \; D_{\text{mask}} \; W_{\mathrm{up}} \in \mathbb{R}^{D\times D}
\]

Applying \(\mathbf{M}\) to \(Y\) yields the sequence-mixed output:
\[
Z = \mathbf{M}\;Y \in \mathbb{R}^{C\times D}
\]

According to Matrix Mixer framework, \textit{MatrixMixer} show the linear Attention class mixer matrices.

In JustDense, each original mixer \(\mathbf{M}\) is replaced with a dense mixer:
\[
\tilde{\mathbf{M}} = f_{\mathrm{Dense}}\bigl(Y\bigr) \in \mathbb{R}^{D\times D}
\]
independent of \(R\) or \(Y\). The resulting output becomes
\[
Z_{\mathrm{dense}} = \tilde{\mathbf{M}}\;Y \in \mathbb{R}^{C\times D}
\]

\subsection{Autoformer}
Let \(\mathbf{X}\in\mathbb{R}^{L\times C}\). We embed:
\begin{equation*}
  f_X(\mathbf{X}) = \mathbf{X}W_V \;+\; W_{\mathrm{pos}},
  \quad
  W_V\in\mathbb{R}^{C\times D},
  \;
  W_{\mathrm{pos}}\in\mathbb{R}^{L\times D}
\end{equation*}
with \(D=H\,P\). For each head \(h\):
\begin{align*}
Q^{(h)} = f_X(\mathbf{X})\,W_Q^{(h)}, \\
K^{(h)} = f_X(\mathbf{X})\,W_K^{(h)}, \\
W_Q^{(h)}, W_K^{(h)} \in \mathbb{R}^{D \times P}
\end{align*}

The autocorrelation is computed as:
\begin{equation*}
    \mathrm{ACorr}^{(h)} = \mathcal{F}^{-1}\bigl(\mathcal{F}(Q^{(h)})\odot\overline{\mathcal{F}(K^{(h)})}\bigr) \in \mathbb{R}^{L}
\end{equation*}
where $\mathcal{F}(\cdot)$ denotes the Fourier transform and $\overline{\mathcal{F}(K^{(h)})}$ is its complex conjugate. For head \(h=1,\dots,H\), the autocorrelation mixer matrix \(\mathbf{M}^{(h)}\in\mathbb{R}^{L\times L}\) has entries
\begin{equation*}
  m_{ij}^{(h)} = \mathrm{ACorr}_{|i - j|}^{(h)}
\end{equation*}

This autocorrelation-based mixer is parameterized by the learnable projection matrices \(W_Q^{(h)}\) and \(W_K^{(h)}\),
which produce the query and key sequences used in the frequency-domain computation.
The head output is:
\begin{equation*}
  \begin{aligned}
    \mathbf{Y}^{(h)}&=\mathbf{M}^{(h)}\,\bigl(f_X(\mathbf{X})\bigr)^{(h)}
  \end{aligned}
\end{equation*}

According to Matrix Mixer framework, \textit{MatrixMixer} show the Toeplitz class mixer matrices.

In JustDense, the autocorrelation mixer is replaced by
\begin{equation*}
  \begin{aligned}
  \tilde{\mathbf{M}}^{(h)}&=f_{\mathrm{Dense}}^{(h)}(\mathbf{X}) \in \mathbb{R}^{L\times L}
  \end{aligned}
\end{equation*}
then
\begin{equation*}
  \begin{aligned}
    \mathbf{Y}_{\mathrm{dense}}^{(h)}&=\tilde{\mathbf{M}}^{(h)}\bigl(f_X(\mathbf{X})\bigr)^{(h)}
  \end{aligned}
\end{equation*}

\subsection{ModernTCN}
Let the original input sequence be \(\mathbf{X}\in\mathbb{R}^{L\times C}\).  
To preserve variable independence, we first apply non‐overlapping patch embedding and downsampling via stacked Conv1d + BN layers:
\begin{align*}
  f_X(\mathbf{X})
  &= \mathrm{DownsampleLayers}(\mathbf{X}) \\
  &= \mathrm{BN}\bigl(\mathrm{Conv1d}(\cdots\,\mathrm{BN}(\mathrm{Conv1d}(\mathbf{X}^\top))\cdots)\bigr) \\
  &\quad\in\;\mathbb{R}^{B \times D \times L'}.
\end{align*}
where each Conv1d has kernel size \(k\), stride \(s\) (for downsampling), and BN is batch normalization; \(D=H\times P\) is the total feature dimension, and \(L'<L\) is the downsampled length.

ModernTCN employs a depth‐wise dilated convolution (DWConv) as its Toeplitz‐like sequence mixer.  For each head \(h=1,\dots,H\), define the mixer matrix \(\mathbf{M}^{(h)}\in\mathbb{R}^{L'\times L'}\) via a learnable kernel \(w^{(h)}\in\mathbb{R}^{K}\), dilation \(d\), and kernel size \(K\):
\begin{equation*}
  m_{ij}^{(h)} =
  \begin{cases}
    w^{(h)}_{\frac{i-j}{d}+1}, 
      & (i - j)\bmod d = 0 \;\wedge\; 1 \le \tfrac{i-j}{d}+1 \le K,\\
    0, & \text{otherwise}.
  \end{cases}
\end{equation*}
Applying this mixer to the \(h\)-th feature slice yields
\[
  \mathbf{Z}^{(h)} 
  = \mathbf{M}^{(h)}\,\bigl(f_X(\mathbf{X})\bigr)^{(h)}
  \;\in\;\mathbb{R}^{L'\times P}
\]
We normalize and activate per channel:
\[
  \mathbf{Z}_{\mathrm{bn}}^{(h)} = \mathrm{BN}\bigl(\mathbf{Z}^{(h)}\bigr), 
  \quad
  \mathbf{Y}^{(h)} = \mathrm{GELU}\bigl(\mathbf{Z}_{\mathrm{bn}}^{(h)}\bigr)
\]

According to Matrix Mixer framework, \textit{MatrixMixer} show the Toeplitz class mixer matrices.

In JustDense, each structured \(\mathbf{M}^{(h)}\) is replaced by
a trainable dense matrix \(\tilde{\mathbf{M}}^{(h)}\in\mathbb{R}^{L'\times L'}\):
\[
  \mathbf{Y}_{\mathrm{dense}}^{(h)}
  = \mathrm{GELU}\Bigl(\mathrm{BN}\bigl(\tilde{\mathbf{M}}^{(h)}\,(f_X(\mathbf{X}))^{(h)}\bigr)\Bigr)
\]
eliminating any convolution‐specific sparsity while preserving the per‐head, per‐variable independence enforced by the downsampling embedding.

\subsection{Mamba}
\label{appendix:mamba}
Let the input sequence be \(\mathbf{X}\in\mathbb{R}^{L\times C}\). We first embed each time step:
\begin{equation*}
f_X(\mathbf{X}) = \mathbf{X}W_{\mathrm{emb}},
\quad
W_{\mathrm{emb}}\in\mathbb{R}^{C\times D}
\end{equation*}
mixer matrix $\mathbf{M}$ has an unstructured parameterization with a full head (H = D).
Mamba constructs a structured mixer matrix using a state‐space formulation with the following learnable parameters:
\begin{gather*}
  A \in \mathbb{R}^{D \times N}, \\
  B = \mathbf{X} W_B \in \mathbb{R}^{L \times N}, \quad W_B \in \mathbb{R}^{C \times N}, \\
  C = \mathbf{X} W_C \in \mathbb{R}^{L \times N}, \quad W_C \in \mathbb{R}^{C \times N}, \\
  \Delta = \mathbf{X} W_\Delta \in \mathbb{R}^{L \times D}, \quad W_\Delta \in \mathbb{R}^{C \times D} \\
  \bar{A} = \exp(\Delta A) \quad \bar{B} = (\Delta A)^{-1} \left( \exp(\Delta A) - I \right) \cdot \Delta B\\
  \bar{A}, \bar{B} \in \mathbb{R}^{L \times D \times N}
\end{gather*}
$A$, $W_B$, $W_C$, and $W_\Delta$ are all learnable parameters.
 For each head \(h=1,\dots,D\), the structured mixer matrix is defined as:
\begin{equation*}
  \begin{aligned}
    m_{ij}^{(h)} = c_i^{(h)\top} \prod_{k=j+1}^{i-1} \bar{A}_k^{(h)} \bar{B}_j^{(h)}
  \end{aligned}
\end{equation*}

The head output is:
\begin{equation*}
  \mathbf{Y}^{(h)}= \mathbf{M}^{(h)}\,\bigl(f_X(\mathbf{X})\bigr)^{(h)}
  \;\in\;\mathbb{R}^{L\times P}
\end{equation*}

According to Matrix Mixer framework, \textit{MatrixMixer} show the semiseparable matrix class mixer matrices.

In JustDense, this structured mixer is replaced by a learnable dense matrix:
\begin{equation*}
  \tilde{\mathbf{M}}^{(h)} = f_{\mathrm{Dense}}^{(h)}(\mathbf{X}) \in \mathbb{R}^{L\times L}
\end{equation*}
then
\begin{equation*}
  \mathbf{Y}_{\mathrm{Dense}}^{(h)} = \tilde{\mathbf{M}}^{(h)}\,\bigl(f_X(\mathbf{X})\bigr)^{(h)}
\end{equation*}

\subsection{S-Mamba}
S-Mamba extends Mamba by incorporating bidirectional sequence mixing, processing both the forward and reversed sequences. It employs two Mamba blocks, defined as:
\begin{equation*}
  \begin{aligned}
  \tilde{\mathbf{X}} &= [\mathbf{X}_L, \mathbf{X}_{L-1}, \ldots, \mathbf{X}_1] \\
  Y &= \mathrm{MambaBlock}(\mathbf{X}) + \mathrm{MambaBlock}(\tilde{\mathbf{X}})
  \;\in\;\mathbb{R}^{L\times D}
  \end{aligned}
\end{equation*}
Here, \(\tilde{\mathbf{X}}\) denotes the reversed sequence.
$MambaBlock$'s MatrixMixer(SSM) is defined in Appendix~\ref{appendix:mamba}. 
In JustDense, we replace both the forward and reverse mixers of S-Mamba with learnable dense matrices, such as mamba.

\section{Evaluation Metrics}
\label{appendix:metrics}

\subsection{Classification}
We adopt classification accuracy, defined as
\[
\mathrm{Accuracy}
=\;\frac{TP + TN}{TP + TN + FP + FN}
\]
where $TP$, $TN$, $FP$, and $FN$ are true positives, true negatives, false positives, and false negatives, respectively.

\subsection{Anomaly Detection} We adopt the F1-Score, the harmonic mean of precision and recall:
\[
\mathrm{F1-Score}
=2\;\frac{\mathrm{Precision}\,\times\,\mathrm{Recall}}{\mathrm{Precision} + \mathrm{Recall}}
\]

\[
\mathrm{Precision}=\frac{TP}{TP + FP}
\]

\[
\mathrm{Recall}=\frac{TP}{TP + FN}
\]

\subsection{Imputation} We adopt the mean squared error (MSE):
\[
\mathrm{MSE}
=\;\frac{1}{T}\sum_{i=1}^{T}\bigl(X_i - \hat{X}_i\bigr)^{2}
\]
where $\hat{X}_i$ and $X_i$ are the imputed and ground-truth values at index $i$, and $T$ is the number of missing entries.

\subsection{Long‐Term Forecasting} We adopt the mean squared error (MSE):
\[
\mathrm{MSE}
=\;\frac{1}{T}\sum_{i=1}^{T}\bigl(X_i - \hat{X}_i\bigr)^{2}
\]
where $\hat{X}_i$ and $X_i$ are the forecast and ground-truth values at time step $i$, and $T$ is the forecast horizon.

\subsection{Short‐Term Forecasting} We adopt the mean absolute scaled error (MASE):
\[
\mathrm{MASE}
=\;\frac{1}{T}\sum_{i=1}^{T}\frac{\bigl|X_i - \hat{X}_i\bigr|}
{\frac{1}{T-1}\sum_{j=2}^{T}\bigl|X_j - X_{j-1}\bigr|}
\]
Here, $T$ denotes the number of forecasted time steps (the forecast horizon), and $i$ runs over each predicted step.
In the denominator, $j$ runs over the in‐sample training indices from $2$ to $T$
to compute the one‐step naive error $|X_j - X_{j-1}|$; thus, scaling the forecast error
by typical historical changes in the series.

\subsection{Attention Map Comparison}
We adopt Peak Signal-to-Noise Ratio (PSNR) and Jensen–Shannon Divergence (JSD):

\[
\text{PSNR}(M,\tilde{M})
=10 \log_{10}\!\Biggl(\frac{M_{\max}^{2}}{\frac{1}{L^{2}}\sum_{i=1}^{L}\sum_{j=1}^{L}\bigl(M_{ij}-\tilde{M}_{ij}\bigr)^{2}}\Biggr)
\]

\begin{equation*}
\begin{aligned}
\mathrm{JSD}(M,\tilde M)
&= \tfrac12 \sum_{i,j=1}^{L} p_{ij}\,\log\!\frac{p_{ij}}{m_{ij}} \\
&\quad+ \tfrac12 \sum_{i,j=1}^{L} q_{ij}\,\log\!\frac{q_{ij}}{m_{ij}}
\end{aligned}
\end{equation*}

where
\begin{description}
  \item[$L$] the matrix dimension.
  \item[$i,j=1,\dots,L$] row and column indices of the matrices.
  \item[$a,b=1,\dots,L$] dummy indices for summation.
  \item[$M_{\max} = \max_{i,j}M_{ij}$] the maximum entry of the original map.
  \item[$p_{ij} = \dfrac{M_{ij}}{\sum_{a,b}M_{ab}}$] normalized distribution of $M$.
  \item[$q_{ij} = \dfrac{\tilde{M}_{ij}}{\sum_{a,b}\tilde{M}_{ab}}$] normalized distribution of $\tilde{M}$.
  \item[$m_{ij} = \tfrac{1}{2}(p_{ij}+q_{ij})$] pointwise mixture distribution.
\end{description}

\section{Rank Analysis of Sequence Mixers}
\label{appendix:rank}

Traditional \textit{attention-based}, \textit{Toeplitz-based} and \textit{semiseparable-based} mixers all reside in
\emph{low-rank} sub-spaces of the full \(L\times L\) matrix space.
This appendix states tight rank bounds for each family, using consistent notation:
\(\mathbf X \in \mathbb{R}^{L\times C}\) input, \(D\) model width, \(P\) head width, \(K\) kernel length, \(N\) state size.

\subsection{Attention Based Mixer}
For single-head self-attention with \(Q,K\!\in\!\mathbb{R}^{L\times P}\),
the score matrix \(S=QK^{\top}/\sqrt{P}\) satisfies \(\operatorname{rank}(S)\le P\)
because \(\operatorname{rank}(QK^{\top})\le P\); row-wise soft-max rescales each row without changing its span; thus, the final attention matrix \(M=\operatorname{softmax}(S)\) obeys \(\operatorname{rank}(M)\le P\) \cite{wang2020linformer}.

\subsection{Toeplitz Based Mixer}
A depth-wise 1-D convolution with kernel length \(K{+}1\) yields a banded Toeplitz mixing matrix \(M\!\in\!\mathbb{R}^{L\times L}\)
whose columns are circular shifts of the kernel.  
Because every column beyond the first \(K{+}1\) is a linear combination of those initial \(K{+}1\) shifts,
the column (and row) space has dimension at most \(K{+}1\), and hence \(\operatorname{rank}(M)\le K{+}1\)~\cite{gray2006toeplitz}.

\subsection{Semiseparable Based Mixer}
For an $N$-state discrete SSM unrolled to length $L$, the resulting mixer
matrix $\mathbf{M}\in\mathbb{R}^{L\times L}$ is $N$-semiseparable—every
strictly lower-triangular block has rank $\le N$~\cite{gu2022efficiently}.
Because its columns lie in the span of the $N$ state responses 
$\{\!\prod_{k=j+1}^{i}\bar{\mathbf A}_k\mathbf b_j\}$; thus, stacking
$\lceil L/N\rceil$ such blocks along the diagonal yields the global bound \(\operatorname{rank}(\mathbf M)\;\le\;N\bigl\lceil L/N\bigr\rceil\)~\cite{gu2022efficiently}.

\subsection{Implicit Low-Rank Bias of Optimisation}
Although a dense matrix has high expressive rank, gradient descent exhibits an \emph{implicit bias}:
for linear or factorised networks it converges to solutions with minimal nuclear norm (hence minimal rank) among all interpolating predictors~\cite{chou2024gradient}.
This demonstrates why our trained dense mixers empirically attain high JSD to structured mixer maps despite their larger ambient rank.

\subsection{Dense Mixer as a High-Rank Super-Set}
Because the space of all \(L \times L\) dense matrices \(\mathbb{R}^{L \times L}\) strictly contains various structured low-rank matrix families
—such as toeplitz, semiseparable matrices—a dense mixer can reproduce their behavior and model higher-rank interactions.  
This inclusion is formalized in the Low Displacement Rank (LDR) Theory,
which generalizes various classic matrix structures through the use of displacement operators~\cite{thomas2018learning}.

\subsection{Summary}
Attention-based, Toeplitz-based and semiseparable-based mixers are all \emph{intrinsically low-rank}; a dense mixer supersets them,
inherits their low-rank structure through optimisation, and captures higher-rank effects when required.

\end{document}